\documentclass[runningheads]{llncs}

\usepackage[T1]{fontenc}
\usepackage{graphicx}
\usepackage{amsmath,amssymb}
\usepackage{booktabs}
\usepackage{multirow}
\usepackage{url}
\usepackage{subcaption}
\usepackage{array}
\usepackage{comment}
\usepackage{url}
\usepackage[hidelinks]{hyperref}
\usepackage{xcolor}

\begin{document}

\title{SynthRCT: Scalable Conditional Deformation Synthesis for Synthetic Repeat CT Generation}
\titlerunning{SynthRCT: Deformation Synthesis for Synthetic Repeat CT Generation}

\author{
Tomas Guija-Valiente\inst{1}\href{https://orcid.org/0009-0000-0911-3317}{\orcidID{0009-0000-0911-3317}} \and
Blanca Rodriguez-Gonzalez\inst{1}\href{https://orcid.org/0009-0007-0982-1293}{\orcidID{0009-0007-0982-1293}} \and
Norberto Malpica\inst{1}\href{https://orcid.org/0000-0003-4618-7459}{\orcidID{0000-0003-4618-7459}}
}

\authorrunning{T. Guija-Valiente et al.}

\institute{
Medical Image Analysis and Biometry Lab, Universidad Rey Juan Carlos, Madrid, Spain\\
\email{tomas.guija@urjc.es}
}

\maketitle
\begingroup
\renewcommand\thefootnote{}
\footnotetext{Preprint version corresponding to the initial submission prior to peer review. A revised version was accepted to the MIART Workshop at MICCAI 2026.}
\addtocounter{footnote}{-1}
\endgroup

\begin{abstract}

In proton therapy, plans are typically optimized on a single planning CT, making robustness evaluation essential under anatomical changes. However, current scenarios often rely on simplified perturbations that poorly capture complex, patient-specific variability. We propose SynthRCT, a scalable conditional generative framework for 3D anatomical deformation synthesis. Based on a conditional variational autoencoder, SynthRCT learns a latent deformation space and decodes sampled latent codes into local stationary velocity fields conditioned on an input anatomy. Local fields are assembled into coherent full-volume transformations, enabling memory-scalable generation for large field-of-view CT data. We validate the approach on respiratory 4DCT data with multiple breathing-phase anatomies per subject. SynthRCT enables patient-specific sampling of plausible anatomical transformations beyond predefined robustness scenarios. Code available at: \url{https://github.com/TomasGuija/SynthRCT}.

\end{abstract}

\section{Introduction}

Proton therapy can concentrate dose around the target with high precision, but its finite beam range makes treatment plans highly sensitive to anatomical deviations between the planning CT and treatment anatomy~\cite{unkelbach2018robust,paganetti2021adaptive}. Inter- and intra-treatment anatomical variations can therefore compromise plans optimized on a single planning CT~\cite{pakela2022management}, making robustness evaluation essential to assess whether target coverage and organ-at-risk sparing are preserved under plausible treatment conditions~\cite{vanherk2002inclusion}. Current workflows often define such conditions through predefined perturbations of the planning image, but these simplified scenarios only approximate the complex, patient-specific anatomical variability observed during treatment. A more flexible alternative is to generate plausible deformations directly from the planning image, enabling evaluation under a richer set of synthetic but realistic anatomical scenarios.

\begin{figure}[t]
\centering
\includegraphics[width=\textwidth]{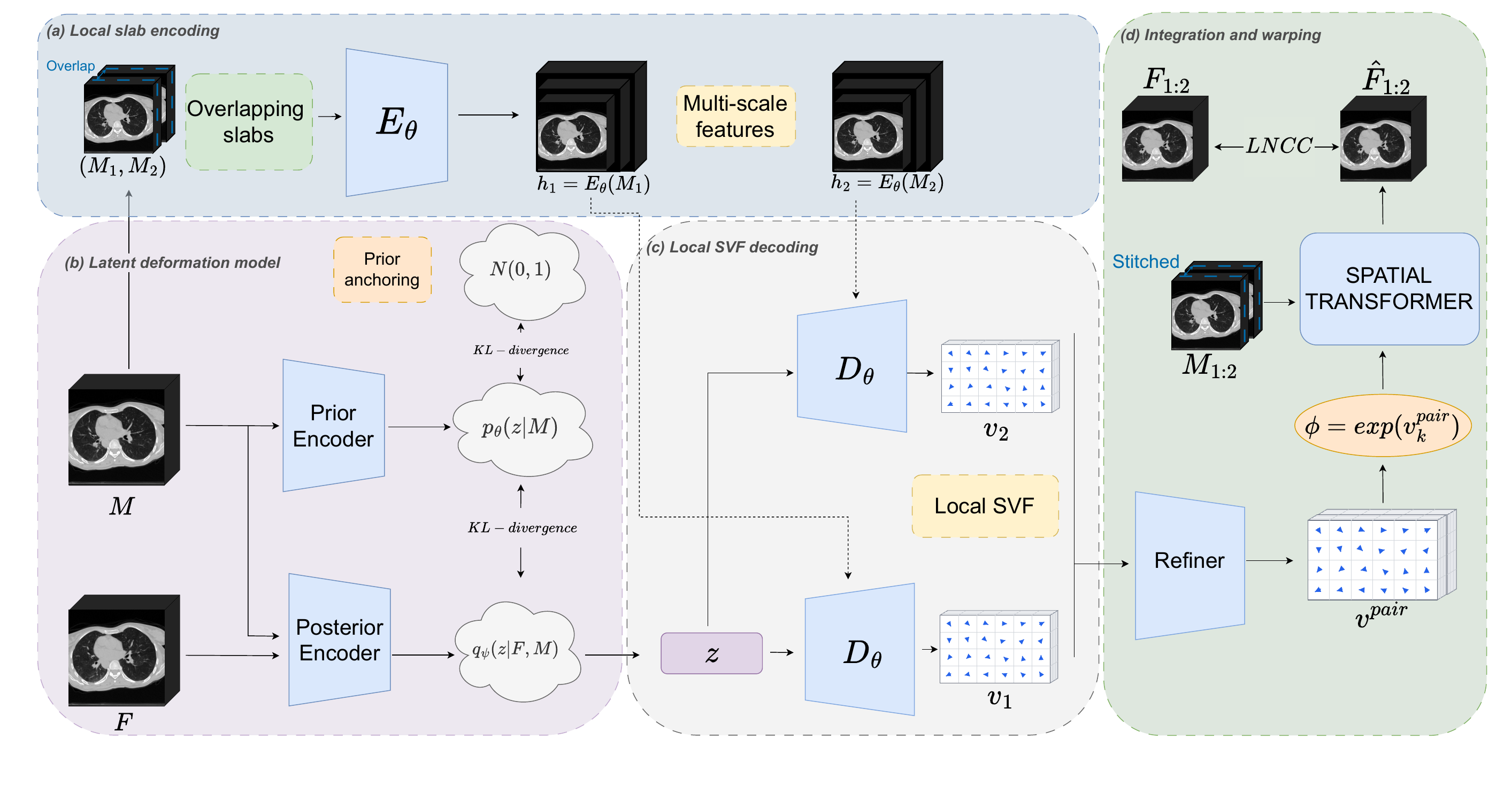}
\caption{
Overview of the proposed SynthRCT framework. A latent deformation code is inferred from a moving--fixed CT pair during training, or sampled from the anatomy-conditioned prior at inference. The code conditions slab-wise SVF decoding, followed by SVF stitching, integration, and differentiable warping to generate synthetic repeat CT anatomies.
}
\label{fig:method_overview}
\end{figure}

Deformable image registration provides the standard framework for estimating transformations between observed images. Widely used methods such as Demons and SyN solve an optimization problem for each image pair \cite{vercauteren2009diffeomorphic,avants2008syn}. Learning-based methods such as VoxelMorph replace iterative optimization with neural networks that predict dense displacement or velocity fields from image pairs \cite{balakrishnan2019voxelmorph,dalca2019probabilistic,mok2020lapirn,chen2022transmorph,skibbe2026patchmorph,wu2024patchfcn}.
However, these methods estimate one transformation between two observed images, rather than sampling multiple plausible transformations from a reference anatomy.

A complementary line of work models anatomical variability, using statistical deformation models \cite{budiarto2011population,szeto2017population,rios2017population}, latent probabilistic models \cite{sohn2015cvae,krebs2019probabilistic,pastor-serrano2023dam}, or deformation-based diffusion models \cite{zheng2026drdm} to sample plausible anatomical changes. Despite these advances, scalable conditional deformation generation for large-field-of-view 3D CT remains challenging, since high-resolution volumetric fields are memory-intensive and local predictions must be combined into coherent full-volume transformations.

We propose SynthRCT, a scalable conditional generative model for 3D anatomical deformation synthesis. SynthRCT learns global latent deformation modes from intra-patient CT pairs and decodes sampled codes into local SVFs over axial slabs, i.e., sub-volumes that cover the full in-plane field of view but only a limited range of axial slices. Local stationary velocity fields are assembled and integrated into a coherent full-volume transformation, reducing memory requirements while preserving high-resolution local deformation modeling, making the approach suitable for large field-of-view CT data.

Our main contributions are a conditional generative framework for patient-specific repeat CT synthesis and a scalable local SVF generation and composition strategy that produces coherent full-volume deformations. We validate the approach through experiments assessing registration accuracy, deformation regularity, latent-space consistency, and landmark-distribution agreement.

\section{Methods}
\subsection{SynthRCT Overview}

\medskip
\noindent\textbf{Conditional latent deformation model}
\par\noindent

Let \(M, F : \Omega \subset \mathbb{R}^{3} \rightarrow \mathbb{R}\) denote two intra-patient CT volumes, where \(M\) is the moving image and \(F\) is the fixed image. SynthRCT learns a conditional distribution of plausible target anatomies \(F\) given an input anatomy \(M\), induced by latent deformation codes \(z \in \mathbb{R}^{d}\):
\begin{equation}
p_{\theta}(F \mid M)
=
\int p_{\theta}(F \mid z, M)\,
p_{\theta}(z \mid M)\,
dz.
\end{equation}

Here, \(p_{\theta}(z \mid M)\) is an anatomy-conditioned prior learned by a CNN encoder that receives the moving image as input, and is regularized toward \(\mathcal{N}(0,I)\) to keep the latent space compact and well behaved. During training, we introduce an approximate posterior distribution \(q_{\psi}(z \mid F,M)\), parameterized by a CNN encoder that observes the moving--fixed pair \((M,F)\). This posterior approximates the latent deformation codes that explain the transformation from \(M\) to \(F\). At inference time, \(F\) is not available; latent codes are therefore sampled from the anatomy-conditioned prior \(p_{\theta}(z \mid M)\).

\medskip
\noindent\textbf{SVF parameterization and integration.}
\par\noindent

SynthRCT parameterizes anatomical change through stationary velocity fields (SVFs). Given an SVF \(v : \Omega \rightarrow \mathbb{R}^{3}\), the transformation is obtained as:
\begin{equation}
    \phi = \exp(v),
\end{equation}
which we approximate using scaling and squaring~\cite{arsigny2006log}. Since a sufficiently smooth velocity field integrates to a smooth invertible transformation, operating in the SVF domain reduces the risk of non-physical foldings compared with directly predicting an unconstrained displacement field.

\medskip
\noindent\textbf{Slab-wise deformation generator.}
\par\noindent

The likelihood \(p_{\theta}(F \mid z, M)\) is defined implicitly by generating an SVF, integrating it into a transformation, and warping the moving image. To scale synthesis to large CT volumes, SynthRCT predicts SVFs locally over axial slabs. An axial slab \(M_{\mathrm{loc}}\) is defined as \(s\) contiguous slices of the moving image \(M\) along the axial direction. The prior and posterior encoders operate on the full-volume anatomies \(M\) and \(F\), so that the latent variable \(z\) captures global deformation modes. In contrast, the deformation generator \(G_{\theta}\) decodes each axial slab \(M_{\mathrm{loc}}\) together with the sampled latent code \(z\) into a local SVF.

The generator \(G_{\theta}\) is implemented as a convolutional U-Net encoder--decoder with skip connections. Its encoder \(E_{\theta}\) extracts multi-resolution features from \(M_{\mathrm{loc}}\), while the decoder \(D_{\theta}\) maps these to a local SVF. The latent code \(z\) is injected into the decoder through Feature-wise Linear Modulation (FiLM)~\cite{perez2018film} allowing the same local anatomy to be deformed according to different sampled deformation modes. This design combines global latent conditioning with local anatomical decoding, enabling coherent respiratory motion modeling while preserving local detail and scaling to large fields of view.

\medskip
\noindent\textbf{Overlap refinement and full-volume assembly.}
\par\noindent

The slab-wise generator produces local SVF predictions, but the final synthetic CT requires a single coherent full-volume deformation. We therefore decode axial slabs with an overlap of \(O\) slices and merge their predictions in the SVF domain before integration. Consider two neighboring overlapping slabs, denoted \(M_1\) and \(M_2\). Both are decoded with the same global latent code \(z\):
\begin{equation}
    v_1 = G_{\theta}(M_1, z),
    \qquad
    v_2 = G_{\theta}(M_2, z).
\end{equation}

Although the shared latent code encourages global consistency, independently decoded slabs may still present small discontinuities at their overlap. To reduce these artifacts, we introduce a convolutional refiner \(S_{\eta}\) that receives the local SVF predictions together with the corresponding moving-image context \(M_{1:2}\), and predicts a residual correction for the overlapping region:
\begin{equation}
    \Delta v_{1:2}
    =
    S_{\eta}
    \left(
        v_1,
        v_2,
        M_{1:2}
    \right).
\end{equation}

The refiner modifies only the overlap. Let \(v_1^{\mathrm{ov}}\) and \(v_2^{\mathrm{ov}}\) denote the restrictions of the two local SVFs to their common overlapping region. The refined overlap is obtained by adding the predicted residual to the average transition:
\begin{equation}
    v_{1:2}^{\mathrm{ov}}
    =
    \frac{1}{2}
    \left(
        v_1^{\mathrm{ov}}
        +
        v_2^{\mathrm{ov}}
    \right)
    +
    \Delta v_{1:2}.
\end{equation}


At inference time, all overlapping slabs are first decoded independently using the same sampled latent code \(z\). The refiner is then applied to each neighboring slab pair, replacing each averaged overlap with its refined counterpart. The resulting regions are assembled into a single full-volume SVF \(v^{\mathrm{full}}\). 

The synthetic repeat CT is finally obtained by warping the input image:
\begin{equation}
\phi^{\mathrm{full}}=\exp(v^{\mathrm{full}}), \qquad \hat F=M\circ\phi^{\mathrm{full}}.
\end{equation}

\subsection{Training Objective}

The model is trained from intra-patient moving--fixed CT pairs \((M,F)\). For each pair, a latent code \(z\) is sampled from the posterior \(q_{\psi}(z \mid F,M)\) and shared across all sampled local regions, encouraging \(z\) to represent global deformation patterns rather than independent local perturbations.

During training, we sample neighboring overlapping slabs from the moving image and decode them with the shared latent code. Their predicted SVFs are refined and assembled into a single local SVF \(v^{\mathrm{pair}}\). This SVF is integrated and used to warp the corresponding moving region:
\begin{equation}
    \phi^{\mathrm{pair}}
    =
    \exp\left(v^{\mathrm{pair}}\right),
    \qquad
    \hat{F}^{\mathrm{pair}}
    =
    M^{\mathrm{pair}}
    \circ
    \phi^{\mathrm{pair}},
\end{equation}
where \(M^{\mathrm{pair}}\) denotes the moving-image region covered by the sampled slab pair.

Image supervision is imposed using local normalized cross-correlation (LNCC). In addition to the final-resolution prediction, the decoder produces intermediate-resolution SVF's used for deep supervision. The similarity loss is therefore
\begin{equation}
    \mathcal{L}_{\mathrm{sim}}
    =
    \mathcal{L}_{\mathrm{LNCC}}^{\mathrm{fine}}
    +
    \lambda_{\mathrm{coarse}}
    \mathcal{L}_{\mathrm{LNCC}}^{\mathrm{coarse}},
\end{equation}
where the coarse term supervises the intermediate decoder outputs after integration and warping at the corresponding resolution.

The latent space is regularized by matching the training posterior to the anatomy-conditioned inference prior, and by anchoring the prior to a standard Gaussian:
\begin{equation}
    \mathcal{L}_{\mathrm{KL}}
    =
    D_{\mathrm{KL}}
    \left(
        q_{\psi}(z \mid F,M)
        \,\|\,
        p_{\theta}(z \mid M)
    \right)
    +
    \alpha
    D_{\mathrm{KL}}
    \left(
        p_{\theta}(z \mid M)
        \,\|\,
        \mathcal{N}(0,I)
    \right).
\end{equation}

The final objective is
\begin{equation}
    \mathcal{L}
    =
    \lambda_{\mathrm{sim}}
    \mathcal{L}_{\mathrm{sim}}
    +
    \beta(t)
    \mathcal{L}_{\mathrm{KL}},
\end{equation}
where \(\lambda_{\mathrm{sim}}\) controls image supervision, \(\lambda_{\mathrm{coarse}}\) weights deep supervision, \(\alpha\) regularizes the conditional prior, and \(\beta(t)\) is a KL warm-up factor.

\section{Experiments}
We evaluate SynthRCT on respiratory 4DCT data, using \(d=32\) latent dimensions and axial slabs of 48 slices with 50\% overlap. Experiments assess image alignment, deformation regularity, latent-space behaviour, and agreement between sampled and observed respiratory landmark distributions. Additional implementation details on the model architecture and training strategy are provided in the publicly available source code repository.

\subsection{Dataset}

Experiments use the DIR4DCT dataset~\cite{castillo2010four,castillo2009framework}, with 10 thoracic 4DCT patients, 10 respiratory phases, and 75 annotated landmark trajectories per subject. Volumes are resampled to 1.5 mm isotropic spacing, normalized to \([0,1]\), and crop/padded around a patient-specific anatomical center to \(256\times256\times207\) voxels. Foreground masks are obtained by intensity thresholding and used to restrict image similarity supervision. Training and evaluation pairs are intra-patient phase pairs separated by at least two respiratory positions to avoid near-identity cases.

\subsection{Registration Accuracy and Deformation Regularity}
\label{sec:registration_accuracy}

Although SynthRCT is not a deterministic registration method, we evaluate posterior-encoded alignment to verify that its latent space reconstructs meaningful respiratory transformations. The posterior encodes the pair \((M,F)\), but the generator receives only the moving slab and latent code \(z\); the fixed image can influence the deformation only through this low-dimensional bottleneck. Therefore, lower alignment accuracy than registration methods such as SyN~\cite{avants2008syn,AVANTS20112033}, Demons~\cite{vercauteren2009diffeomorphic,ITK}, and VoxelMorph~\cite{balakrishnan2019voxelmorph,dalca2019probabilistic} is expected.

\begin{figure}[h]
    \centering

    \begin{subfigure}[t]{0.31\textwidth}
        \centering
        \includegraphics[width=\linewidth]{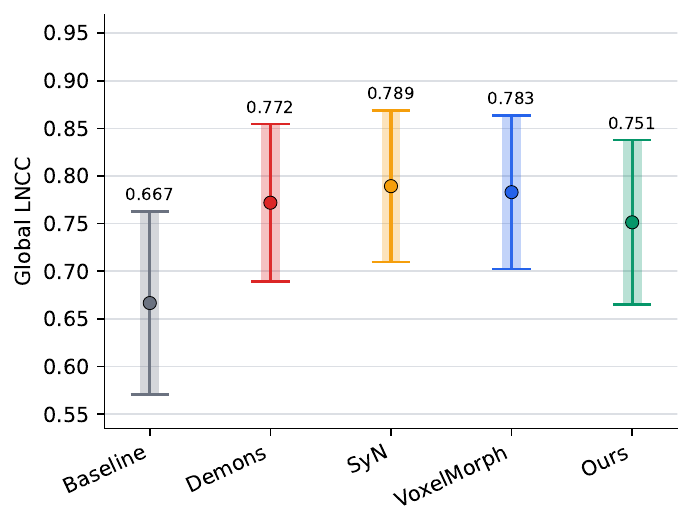}
        \caption{LNCC}
        \label{fig:registration_lncc}
    \end{subfigure}
    \hfill
    \begin{subfigure}[t]{0.31\textwidth}
        \centering
        \includegraphics[width=\linewidth]{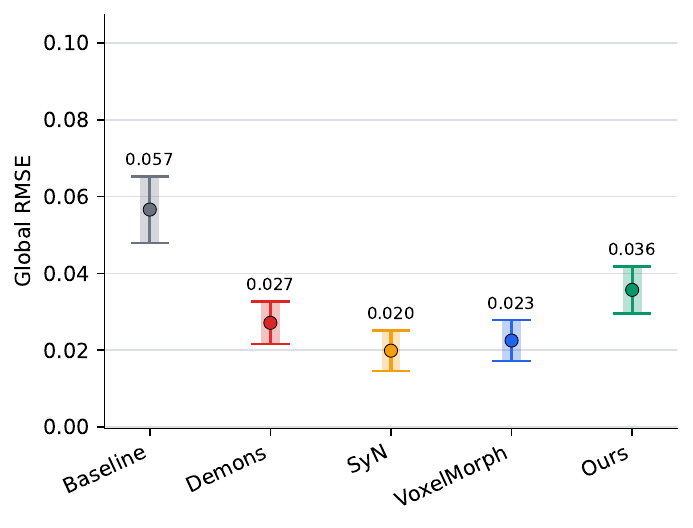}
        \caption{RMSE}
        \label{fig:registration_rmse}
    \end{subfigure}
    \hfill
    \begin{subfigure}[t]{0.31\textwidth}
        \centering
        \includegraphics[width=\linewidth]{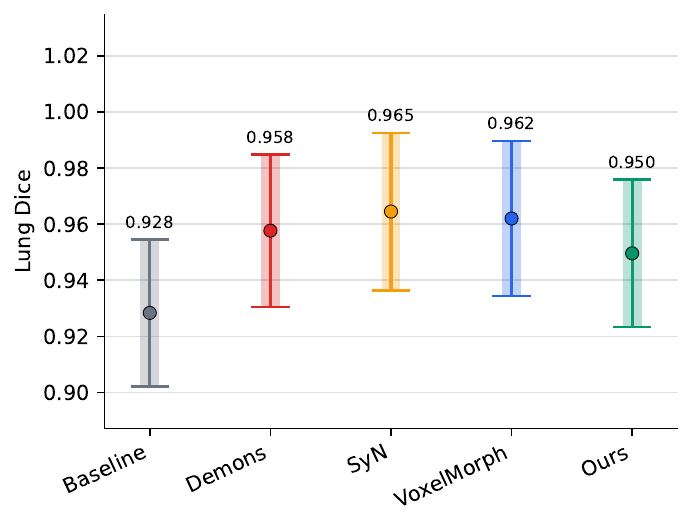}
        \caption{Lung Dice}
        \label{fig:registration_dice}
    \end{subfigure}

    \caption{
    Alignment evaluation on held-out DIR4DCT patients. SynthRCT is compared with the unregistered moving image and deterministic registration methods. Candles show the mean and standard deviation across held-out patients, where each patient-level value is obtained by averaging over all evaluated moving--fixed respiratory phase pairs.
    }
    \label{fig:registration_accuracy}
\end{figure}

We report LNCC, RMSE, and lung Dice overlap (Fig. \ref{fig:registration_accuracy}). SynthRCT improves over the unregistered moving image while remaining below deterministic registration baselines, consistent with its generative objective.

Deformation regularity is assessed through the Jacobian determinant of the transformation \(\phi\). In particular, we measure the folding percentage, defined as the fraction of voxels satisfying \(\det \nabla \phi \leq 0\). Such voxels correspond to local orientation reversals and therefore indicate non-physical topology violations. SynthRCT produced approximately zero folding across the evaluated held-out patients, indicating topology-preserving full-volume transformations.

\subsection{Landmark distribution evaluation}

To evaluate whether sampled deformations reproduce realistic respiratory motion beyond image similarity, we compare real and generated landmark configurations. For each reference phase \(M\), we sample \(N\) latent codes \(z_i \sim p_{\theta}(z \mid M)\), decode them into full-volume SVFs, integrate them, and warp the (L=75) reference landmarks. This yields generated configurations \(\hat{\mathcal{X}}=\{\hat{X}_i\}_{i=1}^{N}\), which are compared with the real respiratory configurations \(\mathcal{X}=\{X_j\}_{j=1}^{R}\). Distances are computed in physical space using the root-mean-square distance.

Table~\ref{tab:landmark_distribution} reports distribution-level metrics between \(\hat{\mathcal{X}}\) and \(\mathcal{X}\). Energy distance \(D_E\) and Wasserstein-1 distance \(W_1\) measure global discrepancy between the real and generated landmark distributions. Coverage \(d_{\mathrm{cov}}\) is the average distance from each real configuration to its nearest generated sample, assessing whether generated samples span the observed respiratory states. Precision \(d_{\mathrm{prec}}\) is the reverse nearest-neighbor distance, measuring whether generated samples remain close to the real distribution. The spread ratio \(\rho_{\mathrm{spread}}\) compares generated and real pairwise variability, with values close to one indicating matched motion diversity.

\begin{table}[h]
    \centering
    \caption{
    Landmark distribution evaluation on the held-out validation patient.
    Distance-based metrics are reported in millimetres.
    }
    \label{tab:landmark_distribution}
    \begin{tabular}{
        l
        @{\hspace{0.8em}}c
        @{\hspace{0.8em}}c
        @{\hspace{0.8em}}c
        @{\hspace{0.8em}}c
        @{\hspace{0.8em}}c
    }
        \toprule
        Case
        & \(D_E\) [mm] \(\downarrow\)
        & \(W_1\) [mm] \(\downarrow\)
        & \(d_{\mathrm{cov}}\) [mm] \(\downarrow\)
        & \(d_{\mathrm{prec}}\) [mm] \(\downarrow\)
        & \(\rho_{\mathrm{spread}}\) \(\rightarrow 1\) \\
        \midrule
        Held-out patient
        & \(1.44\)
        & \(2.61\)
        & \(1.69\)
        & \(1.78\)
        & \(0.98\) \\
        \bottomrule
    \end{tabular}
\end{table}

The generated distribution closely matches the observed respiratory landmark configurations, with millimetre-scale configuration discrepancies.

\subsection{Latent Space Study}
To assess the generative behaviour of SynthRCT, we analyze the learned latent deformation space through interpolation and latent-code statistics.

\begin{figure}[h]
    \centering
    \includegraphics[width=0.6\textwidth]{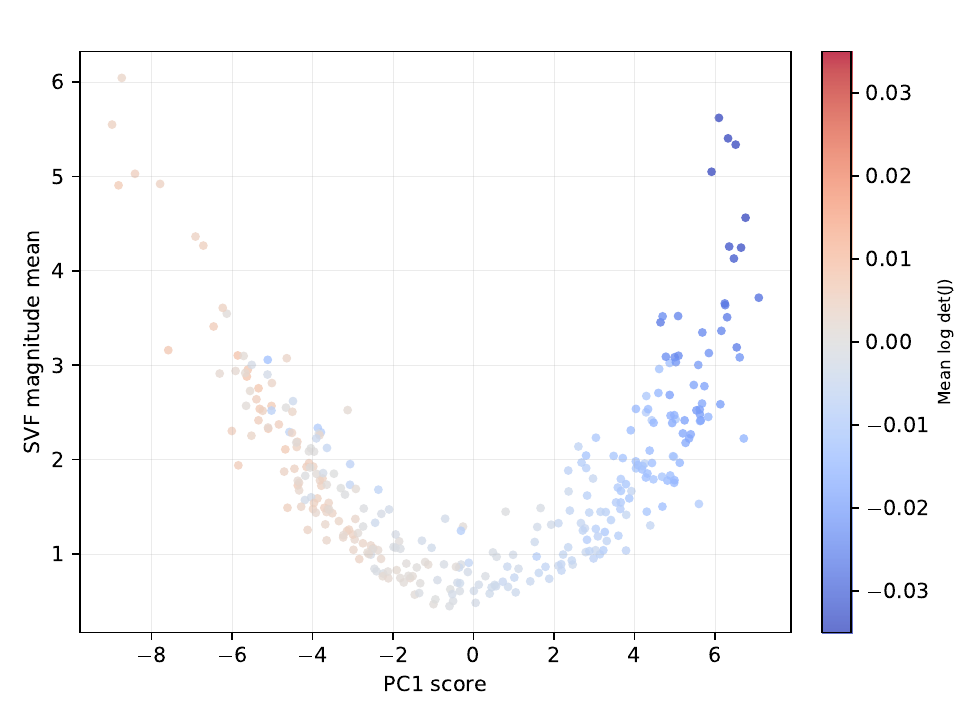}
    \caption{
    PCA analysis of the learned latent deformation space.
    }
    \label{fig:latent_pca_svf_jacobian}
\end{figure}

\begin{figure}[h]
    \centering
    \setlength{\tabcolsep}{2pt}
    \renewcommand{\arraystretch}{0.9}

    \begin{tabular}{
        @{}
        >{\centering\arraybackslash}m{0.08\textwidth}
        *{6}{>{\centering\arraybackslash}m{0.145\textwidth}}
        @{}
    }
        &
        \(\alpha=0.0\) &
        \(\alpha=0.2\) &
        \(\alpha=0.4\) &
        \(\alpha=0.6\) &
        \(\alpha=0.8\) &
        \(\alpha=1.0\) \\

        GT &
        \includegraphics[width=0.145\textwidth]{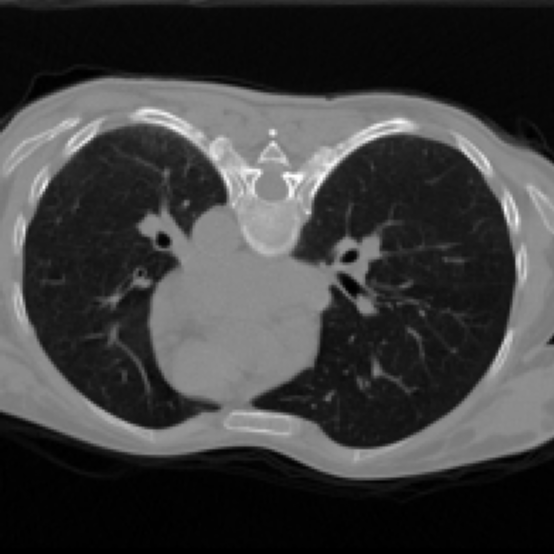} &
        \includegraphics[width=0.145\textwidth]{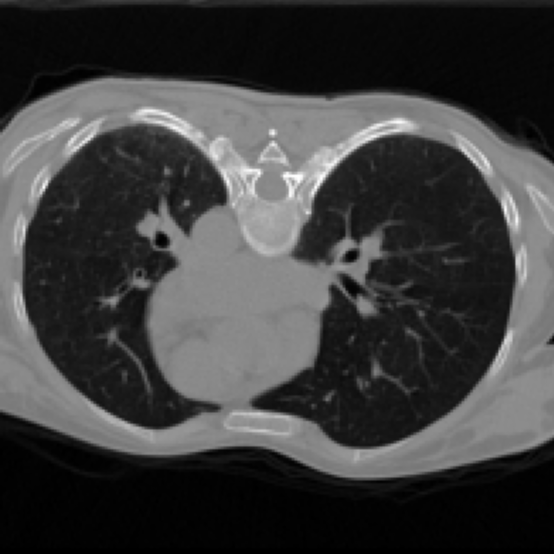} &
        \includegraphics[width=0.145\textwidth]{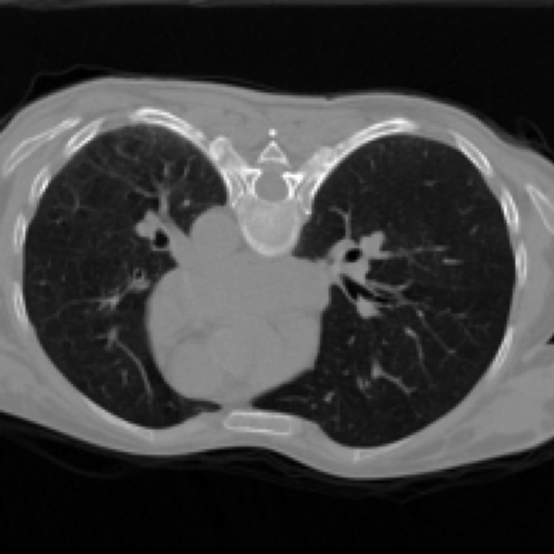} &
        \includegraphics[width=0.145\textwidth]{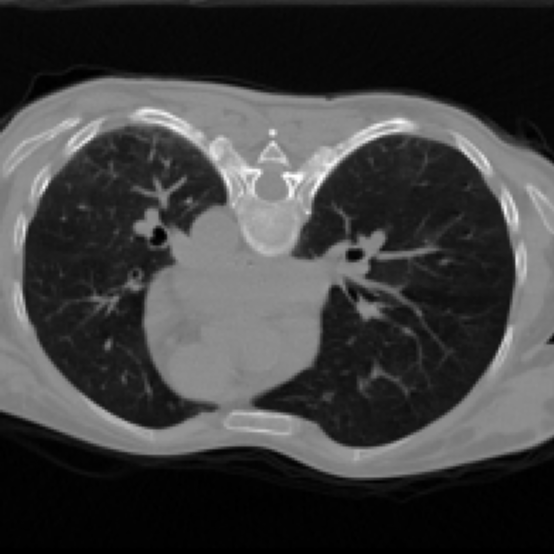} &
        \includegraphics[width=0.145\textwidth]{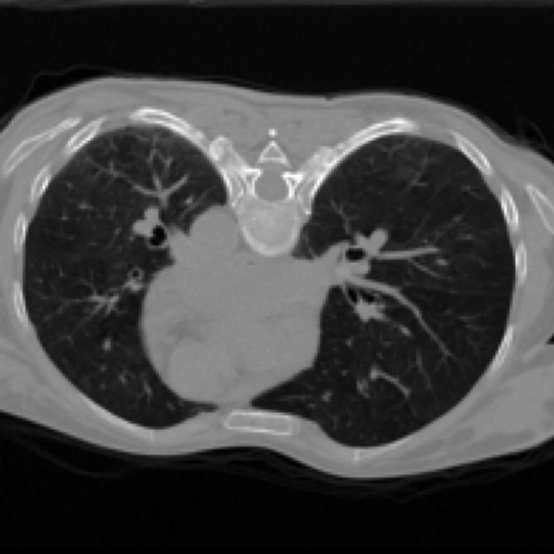} &
        \includegraphics[width=0.145\textwidth]{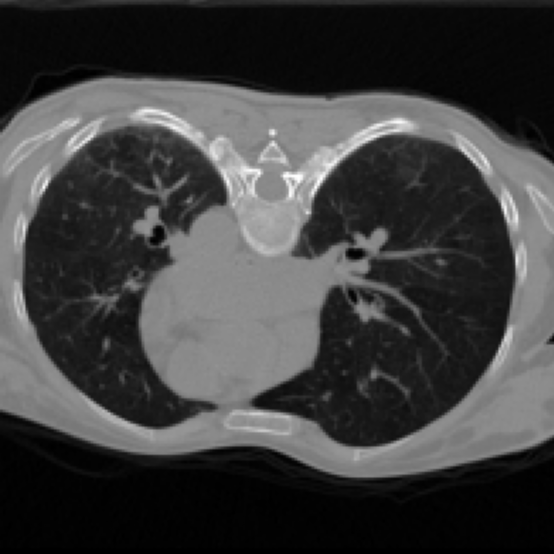} \\

        PRED &
        \includegraphics[width=0.145\textwidth]{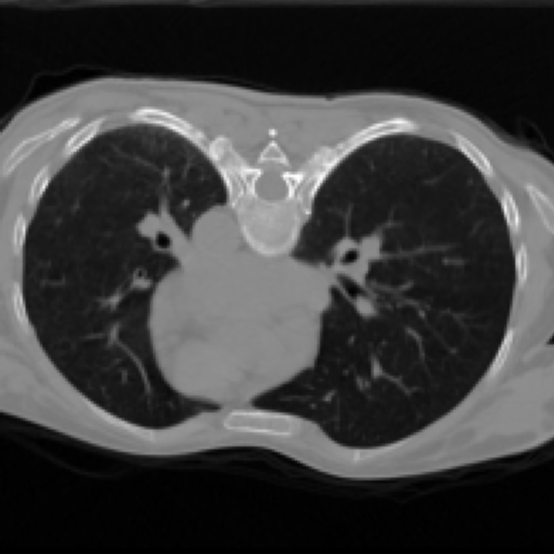} &
        \includegraphics[width=0.145\textwidth]{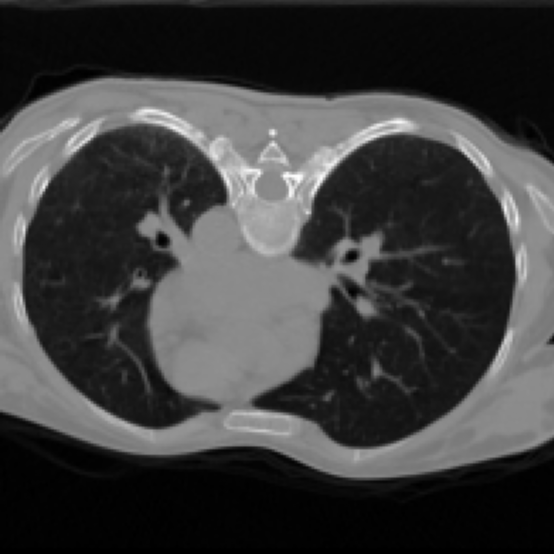} &
        \includegraphics[width=0.145\textwidth]{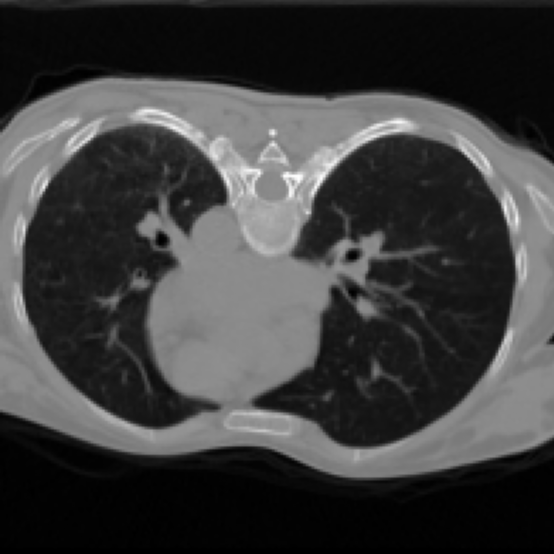} &
        \includegraphics[width=0.145\textwidth]{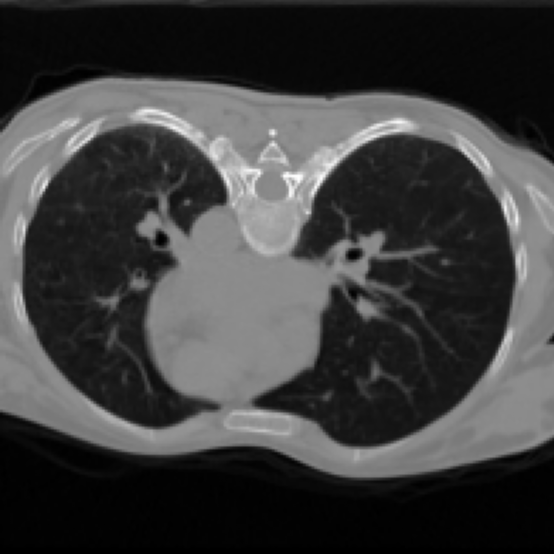} &
        \includegraphics[width=0.145\textwidth]{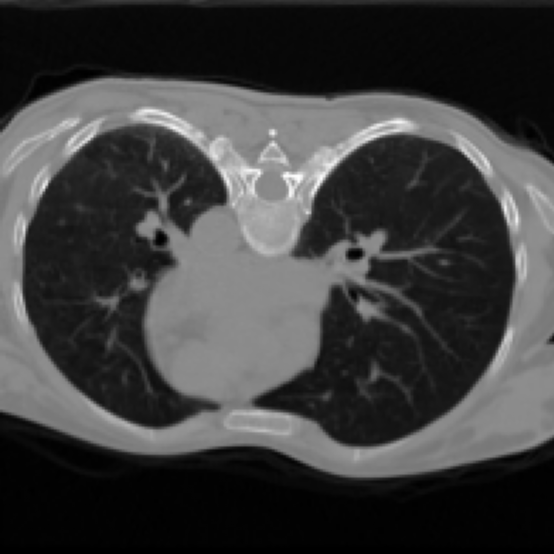} &
        \includegraphics[width=0.145\textwidth]{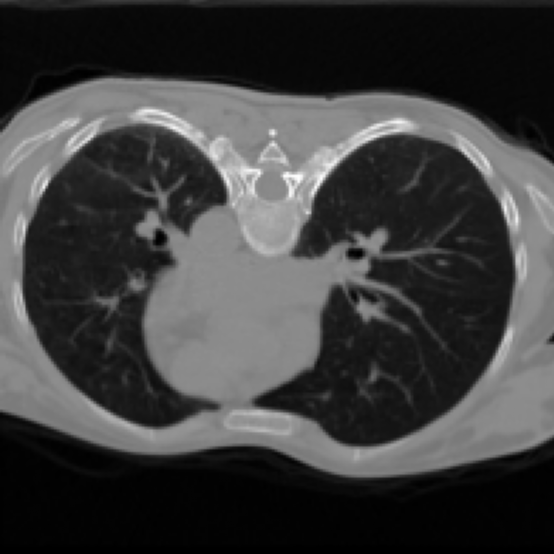} \\

        3D &
        \includegraphics[width=0.145\textwidth]{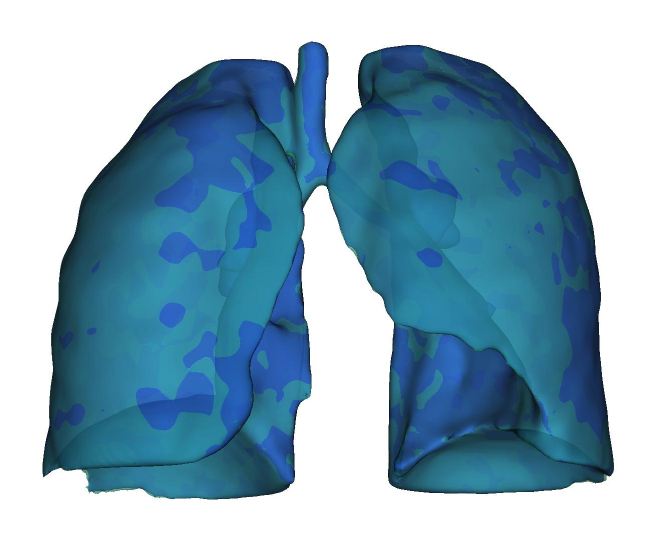} &
        \includegraphics[width=0.145\textwidth]{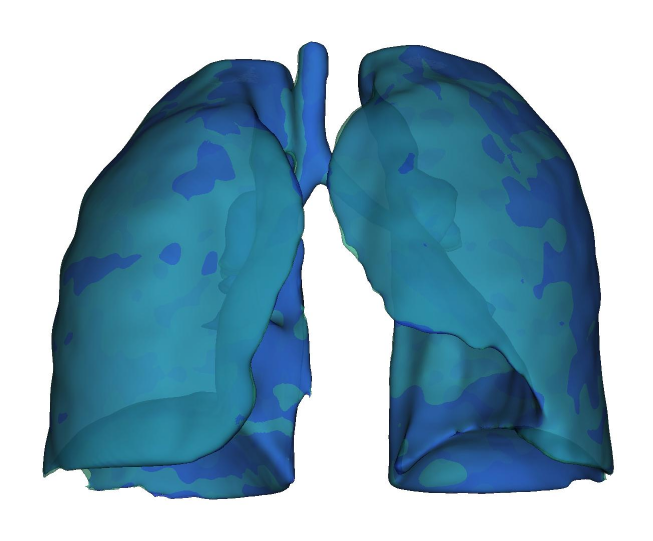} &
        \includegraphics[width=0.145\textwidth]{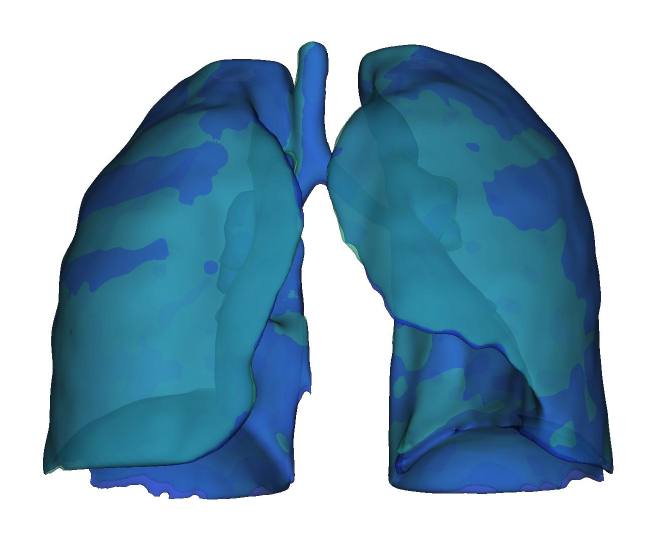} &
        \includegraphics[width=0.145\textwidth]{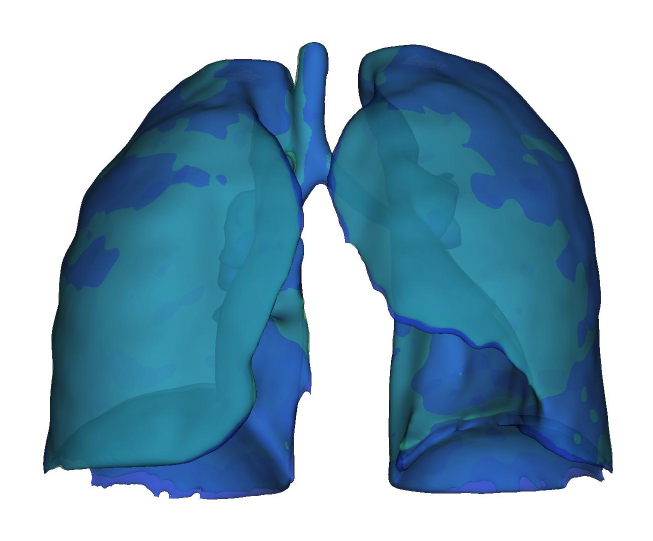} &
        \includegraphics[width=0.145\textwidth]{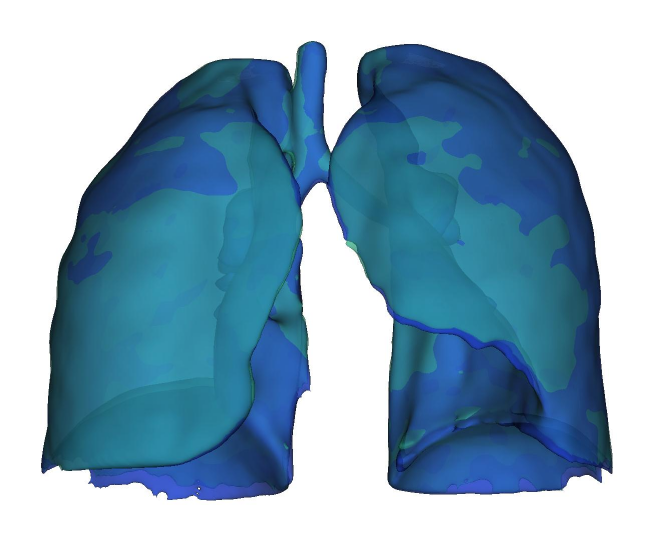} &
        \includegraphics[width=0.145\textwidth]{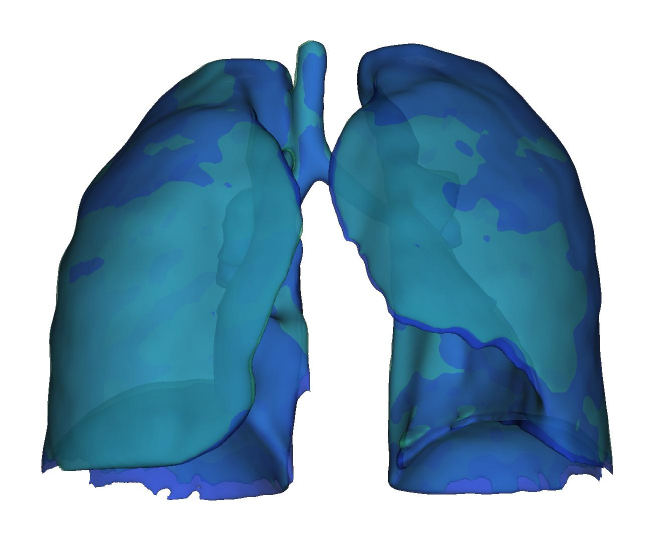} \\

        \(\|v\|\) &
        \includegraphics[width=0.145\textwidth]{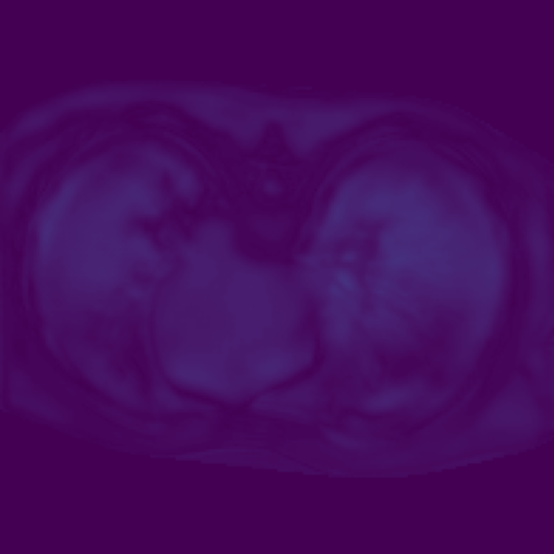} &
        \includegraphics[width=0.145\textwidth]{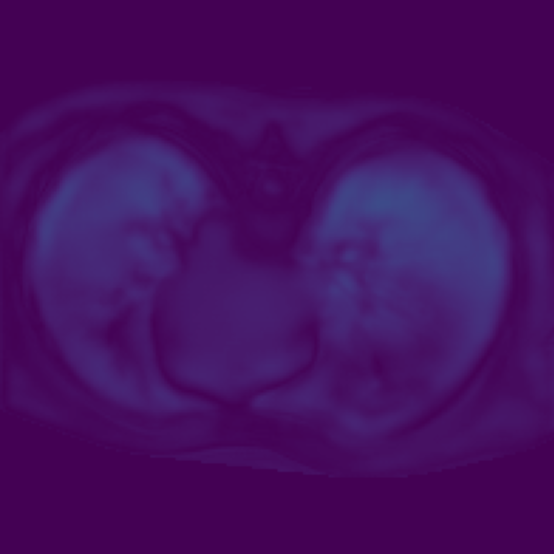} &
        \includegraphics[width=0.145\textwidth]{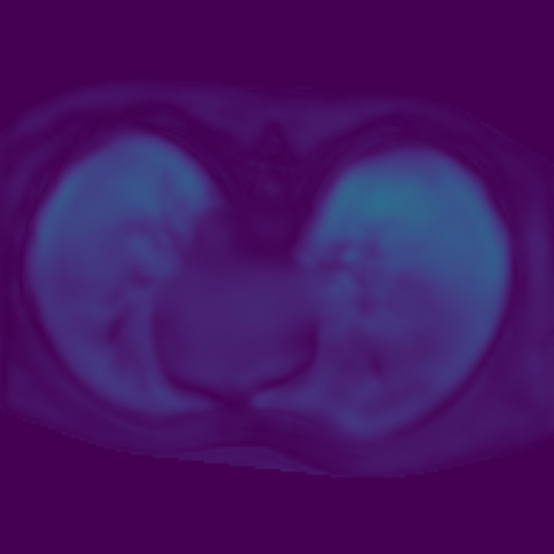} &
        \includegraphics[width=0.145\textwidth]{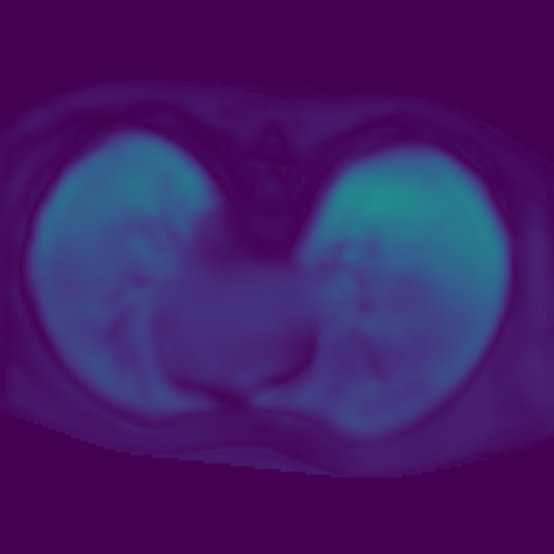} &
        \includegraphics[width=0.145\textwidth]{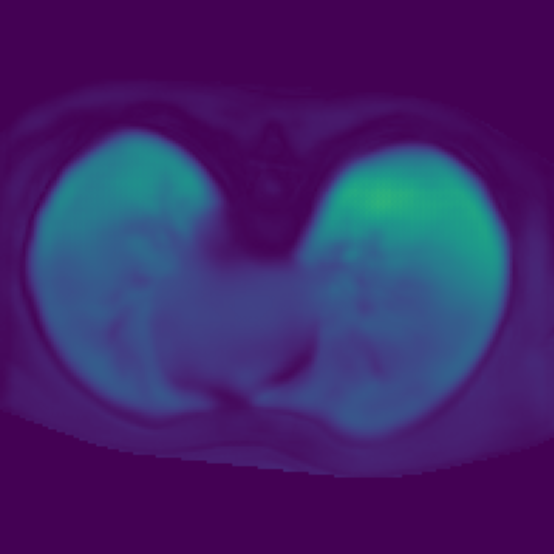} &
        \includegraphics[width=0.145\textwidth]{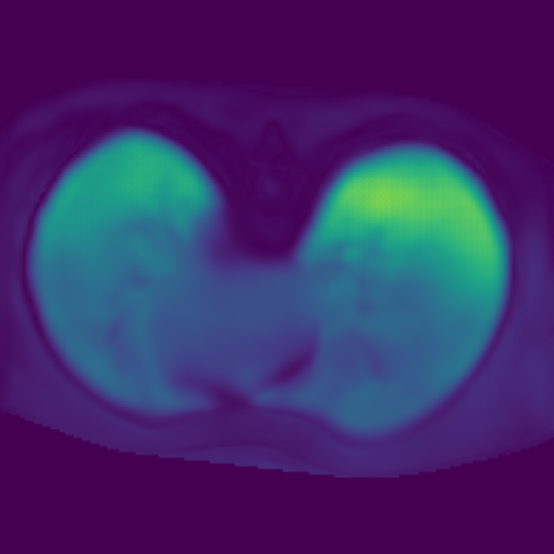}
    \end{tabular}

    \caption{
    Latent-space interpolation on a held-out patient. Columns show linear interpolation between the identity deformation and the largest observed respiratory deformation. Rows show the ground-truth phase, generated warped image, 3D overlay, and SVF magnitude.
    }
    \label{fig:latent_interpolation}
\end{figure}

We test whether linear interpolation in latent space produces smooth respiratory trajectories. For a held-out patient, we interpolate between the identity deformation and the largest observed respiratory deformation, decode the interpolated codes into full-volume SVFs, integrate them, and warp the reference image. As shown in Fig.~\ref{fig:latent_interpolation}, the generated anatomies follow intermediate respiratory phases, with SVF magnitude increasing gradually along the trajectory.

We further analyze latent-space organization by applying PCA to more than 300 posterior latent codes encoded from moving--fixed training pairs. Each code is projected onto the first principal component and compared with deformation-derived statistics.

Fig.~\ref{fig:latent_pca_svf_jacobian} shows that displacement along the main latent direction is associated with increasing deformation magnitude. The color pattern further suggests that opposite directions along this component correspond to different volumetric behaviours, consistent with respiratory expansion and contraction.

\section{Conclusion}

We presented SynthRCT, a conditional probabilistic framework for synthesizing plausible 3D anatomical deformations from a single input CT. The method learns global deformation modes in a latent space while decoding local stationary velocity fields that are assembled before integration into coherent full-volume transformations. This enables sampling and interpolation of anatomical deformations for large field-of-view CT data, reducing peak allocated GPU memory by \(41.6\%\) compared with full-volume decoding. Although pairwise registration is not the main objective, the generated transformations achieved competitive alignment quality while preserving spatial regularity.

The main limitation is the restricted variability of the respiratory 4DCT data, where breathing motion dominates the learned deformation space. Future work should evaluate larger and more diverse datasets to assess whether the latent space can disentangle multiple deformation modes. Additional conditioning signals, such as segmentation masks, could enable controllable deformation synthesis, while more expressive generative models, including flow-matching or diffusion-based approaches, may further improve generation quality.

\begin{credits}
\subsubsection{\ackname} This study has been funded by the MAGERIT-CM project (TEC2024/COM-44), funded by Comunidad de Madrid.
 
\subsubsection{\discintname}
Authors declare no conflict of interests relevant to this research.
\end{credits}


\bibliographystyle{splncs04}
\bibliography{references}

@article{pastor-serrano2023dam,
  author  = {Pastor-Serrano, Oscar and Habraken, Sjoerd and Hoogeman, Mischa and Lathouwers, Danny and Schaart, Dennis and Nomura, Yusuke and Xing, Lei and Perk{\'o}, Zoltan},
  title   = {A probabilistic deep learning model of inter-fraction anatomical variations in radiotherapy},
  journal = {Physics in Medicine and Biology},
  volume  = {68},
  number  = {8},
  pages   = {085018},
  year    = {2023},
  doi     = {10.1088/1361-6560/acc71d}
}

@inproceedings{sohn2015cvae,
  author    = {Sohn, Kihyuk and Lee, Honglak and Yan, Xinchen},
  title     = {Learning Structured Output Representation Using Deep Conditional Generative Models},
  booktitle = {Advances in Neural Information Processing Systems},
  volume    = {28},
  year      = {2015}
}

@inproceedings{perez2018film,
  author    = {Perez, Ethan and Strub, Florian and de Vries, Harm and Dumoulin, Vincent and Courville, Aaron C.},
  title     = {{FiLM}: Visual Reasoning with a General Conditioning Layer},
  booktitle = {Proceedings of the AAAI Conference on Artificial Intelligence},
  year      = {2018}
}

@inproceedings{arsigny2006log,
  author    = {Arsigny, Vincent and Commowick, Olivier and Pennec, Xavier and Ayache, Nicholas},
  title     = {A Log-Euclidean Framework for Statistics on Diffeomorphisms},
  booktitle = {Medical Image Computing and Computer-Assisted Intervention -- MICCAI 2006},
  series    = {Lecture Notes in Computer Science},
  volume    = {4190},
  pages     = {924--931},
  publisher = {Springer},
  address   = {Berlin, Heidelberg},
  year      = {2006},
  doi       = {10.1007/11866565_113}
}

@article{krebs2019probabilistic,
  author  = {Krebs, Julian and Delingette, Herv{\'e} and Mailh{\'e}, Baptiste and Ayache, Nicholas and Mansi, Tommaso},
  title   = {Learning a Probabilistic Model for Diffeomorphic Registration},
  journal = {IEEE Transactions on Medical Imaging},
  volume  = {38},
  number  = {9},
  pages   = {2165--2176},
  year    = {2019},
  doi     = {10.1109/TMI.2019.2897112}
}

@article{avants2008syn,
  author  = {Avants, Brian B. and Epstein, Charles L. and Grossman, Murray and Gee, James C.},
  title   = {Symmetric Diffeomorphic Image Registration with Cross-Correlation: Evaluating Automated Labeling of Elderly and Neurodegenerative Brain},
  journal = {Medical Image Analysis},
  volume  = {12},
  number  = {1},
  pages   = {26--41},
  year    = {2008},
  doi     = {10.1016/j.media.2007.06.004}
}

@article{vercauteren2009diffeomorphic,
  author  = {Vercauteren, Tom and Pennec, Xavier and Perchant, Aymeric and Ayache, Nicholas},
  title   = {Diffeomorphic Demons: Efficient Non-parametric Image Registration},
  journal = {NeuroImage},
  volume  = {45},
  number  = {1},
  pages   = {S61--S72},
  year    = {2009},
  doi     = {10.1016/j.neuroimage.2008.10.040}
}

@article{unkelbach2018robust,
  author  = {Unkelbach, Jan and Paganetti, Harald},
  title   = {Robust Proton Treatment Planning: Physical and Biological Optimization},
  journal = {Seminars in Radiation Oncology},
  volume  = {28},
  number  = {2},
  pages   = {88--96},
  year    = {2018},
  doi     = {10.1016/j.semradonc.2017.11.005}
}

@article{paganetti2021adaptive,
  author  = {Paganetti, Harald and Botas, Paulo and Sharp, Gregory C. and Winey, Brian},
  title   = {Adaptive proton therapy},
  journal = {Physics in Medicine and Biology},
  volume  = {66},
  number  = {22},
  pages   = {22TR01},
  year    = {2021},
  doi     = {10.1088/1361-6560/ac344f}
}

@article{vanherk2002inclusion,
  author  = {van Herk, Marcel and Remeijer, Peter and Lebesque, Joos V.},
  title   = {Inclusion of geometric uncertainties in treatment plan evaluation},
  journal = {International Journal of Radiation Oncology Biology Physics},
  volume  = {52},
  number  = {5},
  pages   = {1407--1422},
  year    = {2002},
  doi     = {10.1016/S0360-3016(01)02805-X}
}

@article{pakela2022management,
  author  = {Pakela, Joshua M. and Knopf, Antje and Dong, Lei and Rucinski, Artur and Zou, Wenchao},
  title   = {Management of Motion and Anatomical Variations in Charged Particle Therapy: Past, Present, and Into the Future},
  journal = {Frontiers in Oncology},
  volume  = {12},
  pages   = {806153},
  year    = {2022},
  doi     = {10.3389/fonc.2022.806153}
}

@article{budiarto2011population,
  author  = {Budiarto, E. and Keijzer, M. and Storchi, P. R. M. and Hoogeman, M. S. and Bondar, L. and Mutanga, T. F. and de Boer, H. C. J. and Heemink, A. W.},
  title   = {A population-based model to describe geometrical uncertainties in radiotherapy: Applied to prostate cases},
  journal = {Physics in Medicine and Biology},
  volume  = {56},
  number  = {4},
  pages   = {1045--1061},
  year    = {2011},
  doi     = {10.1088/0031-9155/56/4/011}
}

@article{balakrishnan2019voxelmorph,
  author  = {Balakrishnan, Guha and Zhao, Amy and Sabuncu, Mert R. and Guttag, John and Dalca, Adrian V.},
  title   = {{VoxelMorph}: A Learning Framework for Deformable Medical Image Registration},
  journal = {IEEE Transactions on Medical Imaging},
  volume  = {38},
  number  = {8},
  pages   = {1788--1800},
  year    = {2019},
  doi     = {10.1109/TMI.2019.2897538}
}

@article{szeto2017population,
  author  = {Szeto, Y. Z. and Witte, M. G. and van Herk, M. and Sonke, J.-J.},
  title   = {A population based statistical model for daily geometric variations in the thorax},
  journal = {Radiotherapy and Oncology},
  volume  = {123},
  number  = {1},
  pages   = {99--105},
  year    = {2017},
  doi     = {10.1016/j.radonc.2017.02.012}
}

@article{rios2017population,
  author  = {Rios, Roberto and de Crevoisier, Renaud and Ospina, Juan David and Commandeur, Fr{\'e}d{\'e}ric and Lafond, Caroline and Simon, Antoine and Haigron, Pascal and Espinosa, Jorge and Acosta, Oscar},
  title   = {Population model of bladder motion and deformation based on dominant eigenmodes and mixed-effects models in prostate cancer radiotherapy},
  journal = {Medical Image Analysis},
  volume  = {38},
  pages   = {133--149},
  year    = {2017},
  doi     = {10.1016/j.media.2017.03.001}
}

@article{castillo2010four,
  author  = {Castillo, Edward and Castillo, Richard and Martinez, Jorge and Shenoy, M. and Guerrero, Thomas},
  title   = {Four-dimensional deformable image registration using trajectory modeling},
  journal = {Physics in Medicine and Biology},
  volume  = {55},
  number  = {1},
  pages   = {305--327},
  year    = {2010},
  doi     = {10.1088/0031-9155/55/1/018}
}

@article{castillo2009framework,
  author  = {Castillo, Richard and Castillo, Edward and Guerra, Rafael and Johnson, Victoria E. and McPhail, Thomas and Garg, Amit K. and Guerrero, Thomas},
  title   = {A framework for evaluation of deformable image registration spatial accuracy using large landmark point sets},
  journal = {Physics in Medicine and Biology},
  volume  = {54},
  number  = {7},
  pages   = {1849--1870},
  year    = {2009},
  doi     = {10.1088/0031-9155/54/7/001}
}

@article{skibbe2026patchmorph,
  author  = {Skibbe, Henrik and Byra, Micha{\l} and Watakabe, Akiya and Yamamori, Tetsuo and Reisert, Marco},
  title   = {Memory Efficient Training for 3D Brain Image Registration Networks Using {PatchMorph}},
  journal = {Scientific Reports},
  volume  = {16},
  pages   = {14386},
  year    = {2026},
  doi     = {10.1038/s41598-026-44858-x}
}

@inproceedings{wu2024patchfcn,
  author    = {Wu, Junde and Zhou, Shihua and Lin, Li and Wang, Xiaoyu and Tan, Wenkai},
  title     = {Fast Diffeomorphic Image Registration Using Patch Based Fully Convolutional Networks},
  booktitle = {Proceedings of the 46th Annual International Conference of the IEEE Engineering in Medicine and Biology Society},
  pages     = {1--4},
  year      = {2024},
  doi       = {10.1109/EMBC53108.2024.10781975}
}

@article{dalca2019probabilistic,
  author  = {Dalca, Adrian V. and Balakrishnan, Guha and Guttag, John and Sabuncu, Mert R.},
  title   = {Unsupervised Learning of Probabilistic Diffeomorphic Registration for Images and Surfaces},
  journal = {Medical Image Analysis},
  volume  = {57},
  pages   = {226--236},
  year    = {2019},
  doi     = {10.1016/j.media.2019.07.006}
}

@inproceedings{mok2020lapirn,
  author    = {Mok, Tony C. W. and Chung, Albert C. S.},
  title     = {Large Deformation Diffeomorphic Image Registration with Laplacian Pyramid Networks},
  booktitle = {Medical Image Computing and Computer Assisted Intervention -- MICCAI 2020},
  series    = {Lecture Notes in Computer Science},
  volume    = {12263},
  pages     = {211--221},
  publisher = {Springer},
  address   = {Cham},
  year      = {2020},
  doi       = {10.1007/978-3-030-59716-0_21}
}

@article{chen2022transmorph,
  author  = {Chen, Junyu and Frey, Eric C. and He, Yufan and Segars, W. Paul and Li, Ye and Du, Yong},
  title   = {{TransMorph}: Transformer for Unsupervised Medical Image Registration},
  journal = {Medical Image Analysis},
  volume  = {82},
  pages   = {102615},
  year    = {2022},
  doi     = {10.1016/j.media.2022.102615}
}

@article{zheng2026drdm,
  author  = {Zheng, Jia-Qi and Mo, Yiming and Sun, Yucheng and Li, Jing and Wu, Fuping and Wang, Ziyang and Vincent, Tristan and Papie{\.z}, Bart{\l}omiej W.},
  title   = {Deformation-Recovery Diffusion Model ({DRDM}): Instance Deformation for Image Manipulation and Synthesis},
  journal = {Medical Image Analysis},
  volume  = {110},
  pages   = {103987},
  year    = {2026},
  doi     = {10.1016/j.media.2026.103987}
}

@article{AVANTS20112033,
title = {A reproducible evaluation of ANTs similarity metric performance in brain image registration},
journal = {NeuroImage},
volume = {54},
number = {3},
pages = {2033-2044},
year = {2011},
issn = {1053-8119},
doi = {https://doi.org/10.1016/j.neuroimage.2010.09.025},
url = {https://www.sciencedirect.com/science/article/pii/S1053811910012061},
author = {Brian B. Avants and Nicholas J. Tustison and Gang Song and Philip A. Cook and Arno Klein and James C. Gee}
}

@article{ITK,
author = {Lowekamp, Bradley and Chen, David and Ibanez, Luis and Blezek, Daniel},
year = {2013},
month = {12},
pages = {45},
title = {The design of simpleITK},
volume = {7},
journal = {Frontiers in neuroinformatics},
doi = {10.3389/fninf.2013.00045}
}

\end{document}